\documentclass{article}
\usepackage{ijcai19}
\usepackage{times}
\usepackage{latexsym}
\usepackage{times}
\usepackage{soul}
\usepackage{url}
\usepackage[hidelinks]{hyperref}
\usepackage[utf8]{inputenc}
\usepackage[small]{caption}
\usepackage{graphicx}
\usepackage{amsmath}
\usepackage{booktabs}
\usepackage{algorithm}
\usepackage{algorithmic}
\title{Improving Question Answering by Commonsense-Based Pre-Training}

\author{Wanjun Zhong$^\S$\thanks{\ \ Work is done during internship at Microsoft Research Asia.}\ \ , Duyu Tang$^\ddag$, Nan Duan$^\ddag$, Ming Zhou$^\ddag$, Jiahai Wang$^\S$ Jian Yin$^\S$,  \\
	$^\S$Sun Yat-Sen University, GuangZhou, China\\
	$^\ddag$Microsoft Research Asia, Beijing, China \\
	{\small \tt zhongwj25@mail2.sysu.edu.cn} \\
	{\small \tt \{dutang,nanduan,mingzhou\}@microsoft.com}\\
	{\small \tt \{wangjiah,issjyin\}@mail.sysu.edu.cn}}

\begin{document}
\maketitle
\newcommand{\newcite}[1]
{\citeauthor{#1}~\shortcite{#1}}
\begin{abstract}
	Although neural network approaches achieve remarkable success on a variety of NLP tasks, many of them struggle to answer questions that require commonsense knowledge.
	We believe the main reason is the lack of commonsense \mbox{connections} between concepts.
	To remedy this, we provide a simple and effective method that leverages external commonsense knowledge base such as ConceptNet.
	We pre-train direct and indirect relational functions between concepts, and 
	show that these pre-trained functions could be easily added to existing neural network models.
	Results show that incorporating commonsense-based function improves the baseline on three question answering tasks that require commonsense reasoning.
	Further analysis shows that our system \mbox{discovers} and leverages useful evidence from an external commonsense knowledge base, which is missing in existing neural network models and help derive the correct answer.
\end{abstract}

\section{Introduction}
\begin{figure}[h]
	\centering
	\includegraphics[width=.48\textwidth]{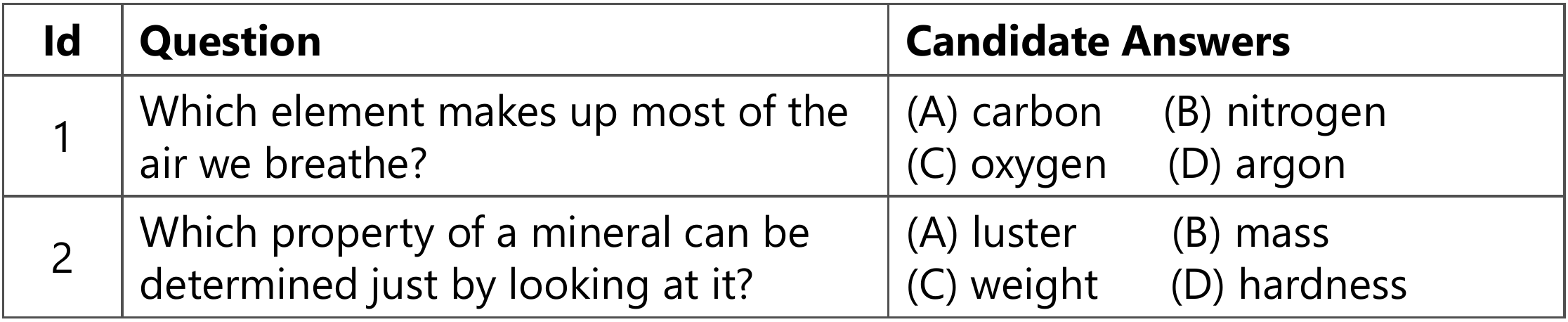}
	\caption{Examples from ARC  that require commonsense knowledge and reasoning.}
	\label{fig:intro-example}
\end{figure}

Commonsense reasoning is a major challenge for question answering  \cite{levesque2011winograd,clark2018think,ostermann2018semeval,boratko2018systematic}. 
Take Figure \ref{fig:intro-example} as an example. 
Answering both questions requires a natural language understanding system that has the ability of reasoning based on commonsense knowledge about the world.
Although neural network approaches have achieved promising performance when supplied with a large number of supervised training instances, even surpassing human-level exact match accuracy on the Stanford Question
Answering Dataset (SQuAD) benchmark \cite{rajpurkar2016squad},
it has been shown that  
existing systems lack true language understanding and reasoning capabilities \cite{jia2017adversarial}, which are crucial to commonsense reasoning. 
Moreover, although it is easy for humans to answer the questions mentioned above based on their knowledge about the world, it is a great  
challenge for machines when there is limited training data.

In this paper, we leverage external commonsense knowledge, such as ConceptNet \cite{speer2012representing}, to improve the commonsense reasoning capability of a question answering (QA) system. 
We believe that a desirable way is to pre-train a generic model from external commonsense knowledge about the world, with the following advantages.
First, such a model has a broader coverage of the concepts/entities and can access rich contexts from the relational knowledge graph.
Second, the ability of commonsense reasoning is not limited to the number of training instances and the coverage of reasoning types in the end tasks.
Third, it is convenient to build a hybrid system that preserves the semantic matching ability of the existing QA system, which might be a neural network-based model, and further integrates a generic model to improve model's capability of commonsense reasoning.

We believe that the main reason why the majority of existing methods lack the commonsense reasoning ability is the absence of 
connections between concepts\footnote{In this work, concepts are words and phrases that can be extracted from natural language text \cite{speer2012representing}.}.
These connections could be divided into direct and indirect ones. 
Below is an example sampled from ConceptNet. 
\begin{figure}[h]
	\centering
	\includegraphics[width=.48\textwidth]{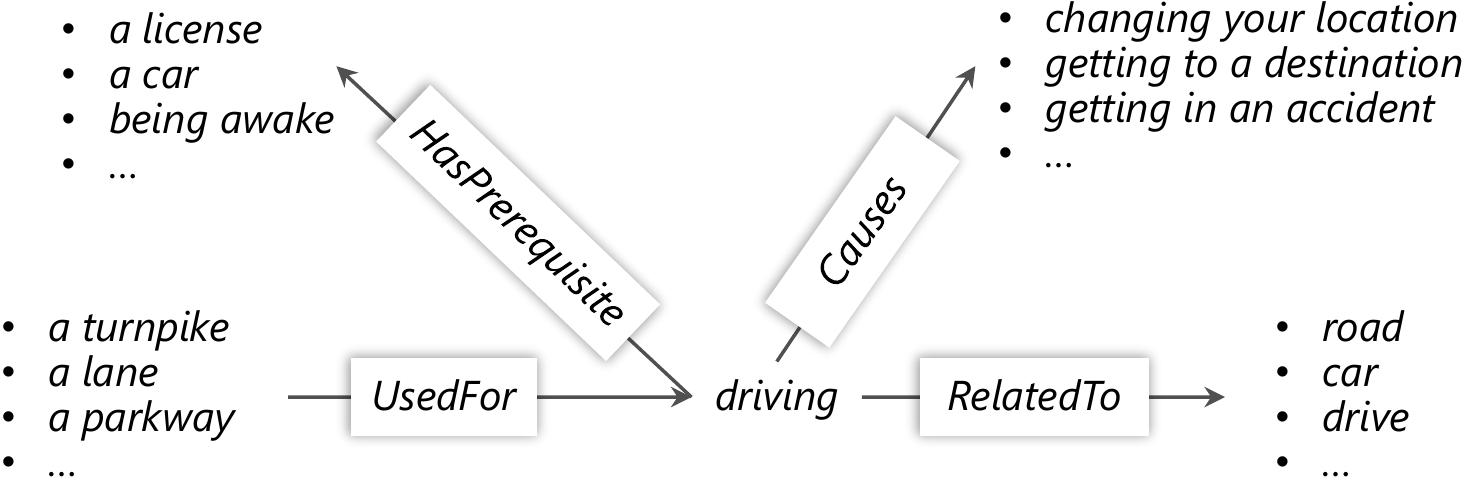}
	\caption{A sampled subgraph from ConceptNet with ``\textit{driving}'' as the central word.}
	\label{fig:kb-sample}
\end{figure}
In this case, \{``\textit{driving}'', ``\textit{a license}''\} forms a direct connection whose relation is ``\textit{HasPrerequisite}''.
\{``\textit{driving}'', ``\textit{road}''\} also forms a direct connection.
Moreover, there are indirect connections here such as \{``\textit{a car}'', ``\textit{getting to a destination}''\}, which are connected by a pivot concept ``\textit{driving}''. 
Based on this, people can learn two functions to measure direct and indirect connections between every pair of concepts. 
These functions could be easily combined with existing QA systems.

We take three question answering tasks \cite{clark2018think,ostermann2018semeval,mihaylov2018can} that require commonsense reasoning as the testbeds. These tasks take a question and optionally a context\footnote{The definitions of contexts in these tasks are slightly different and we will describe the details in the next section.} as input, and select an answer from a set of candidate answers. 
We believe that understanding and answering the question requires knowledge of both words and the world \cite{hirsch2003reading}.
Thus, we implement document-based neural network based baselines and use the same way to improve the baseline systems with our commonsense-based pre-trained models. 
Results show that incorporating pre-trained models brings improvements on these three tasks and improve model's ability to 
discover useful evidence from an external commonsense knowledge base.
The first contribution of our work is that we present a simple yet effective way to pre-train commonsense-based functions to capture the semantic relationships between concepts. The pre-training model can be easily incorporated into other tasks requiring commonsense reasoning.
Secondly, we demonstrate that incorporating the pre-trained model improves strong baselines on three multi-choice question answering datasets.

\section{Tasks and Datasets}
In this work, we focus on integrating commonsense knowledge as a source of supportive information into the question answering task. 
To verify the effectiveness of our approach, we use three multiple-choice question answering tasks that require commonsense reasoning as our testbeds. 

Given a question of length $M$ and optionally a supporting passage of length $N$, both tasks are to \mbox{predict} the correct answer from a set of candidate answers.
The difference between these tasks is the definition of the supporting passage which will be described later in this section.
Systems are expected to select the correct answer from multiple candidate answers by reasoning out the question and the supporting passage.
Following previous studies, we regard the problem as a ranking task. 
At the test time, the model should return the answer with the highest score as the prediction.

The \textbf{first} task comes from SemEval 2018 Task 11\footnote{\url{https://competitions.codalab.org/competitions/17184}} \cite{ostermann2018semeval}, which aims to 
evaluate a system's ability to perform commonsense reasoning in question answering.  
The dataset describes events about daily activities. 
For each question, the supporting passage is a specific document given as a part of the input, and the number of candidate answers is two.
Answering the substantial number of questions presented in this dataset requires inference from commonsense knowledge of diverse scenarios, which are beyond the facts explicitly mentioned in the document.

The \textbf{second} task we focus on is ARC, short for AI2 Reasoning Challenge, proposed by \newcite{clark2018think}\footnote{\url{http://data.allenai.org/arc/arc-corpus/}}.
The ARC Dataset consists of a collection of scientific questions and a large scientific text corpus containing a large number of science facts. 
Each question has multiple candidate answers (mostly 4-way multiple candidate answers). 
The dataset is separated into an easy set and a challenging set.
The Challenging Set contains only difficult, grade-school questions including questions answered incorrectly by both a retrieval-based algorithm and a word co-occurrence algorithm, and have acquired strong reasoning ability of commonsense knowledge or other reasoning procedure \cite{boratko2018systematic}. 
Figure \ref{fig:intro-example} shows two examples which need to be solved by common sense. 
We only use the challenge set.

The \textbf{third} dataset we use in the experiment is OpenBook QA\footnote{\url{http://data.allenai.org/OpenBookQA}}, which calls for exploring the knowledge from an open book fact and commonsense knowledge from other sources. \cite{mihaylov2018can}. 
The dataset consists of 5,957 multiple-choice questions (4,957/500/500 for training/validation/test) and a set of 1,326 facts about elementary level science.

\section{Commonsense Knowledge}
This section describes the commonsense knowledge base we investigate in our experiment. 
We use ConceptNet\footnote{\url{http://conceptnet.io/}} \cite{speer2012representing}, one of the most widely used commonsense knowledge bases.
Our approach is generic and could also be applied to other commonsense knowledge bases such as WebChild \cite{tandon2017webchild}, which we leave as future work.
ConceptNet is a semantic network that represents the large sets of words and phrases and the commonsense relationships between them. 
It contains 657,637 instances and 39 types of relationships. Each instance in ConceptNet can be generally described as a triple $r_i=(subject,relation,object) $.
For example, the ``\textit{IsA}'' relation (e.g. ``\textit{car}'', ``\textit{IsA}'', ``\textit{vehicle}'') means that ``\textit{XX is a kind of YY}''; the ``\textit{Causes}'' relation (e.g. ``\textit{car}'', ``\textit{Causes}'', ``\textit{pollution}'') means that ``\textit{the effect of XX is YY}''; the ``\textit{CapableOf}'' relation (e.g. ``\textit{car}'', ``\textit{CapableOf}'', ``\textit{go fast}'') means that ``\textit{XX can YY}'', etc. More relations and explanations could be found at \newcite{speer2012representing}.

\section{Approach Overview}

In this section, we give an overview of our framework to show the basic idea of solving the commonsense reasoning problem. Details of each component will be described in the following sections.

At the top of our framework, we suggest that we should select the candidate answer with the highest probability (highest score) as our final prediction. So we can tackle this problem by designing a scoring function that captures the evidence mentioned in the passage and retrieved from the commonsense knowledge base. 
\begin{figure}[h]
	\centering
	\includegraphics[width=.45\textwidth]{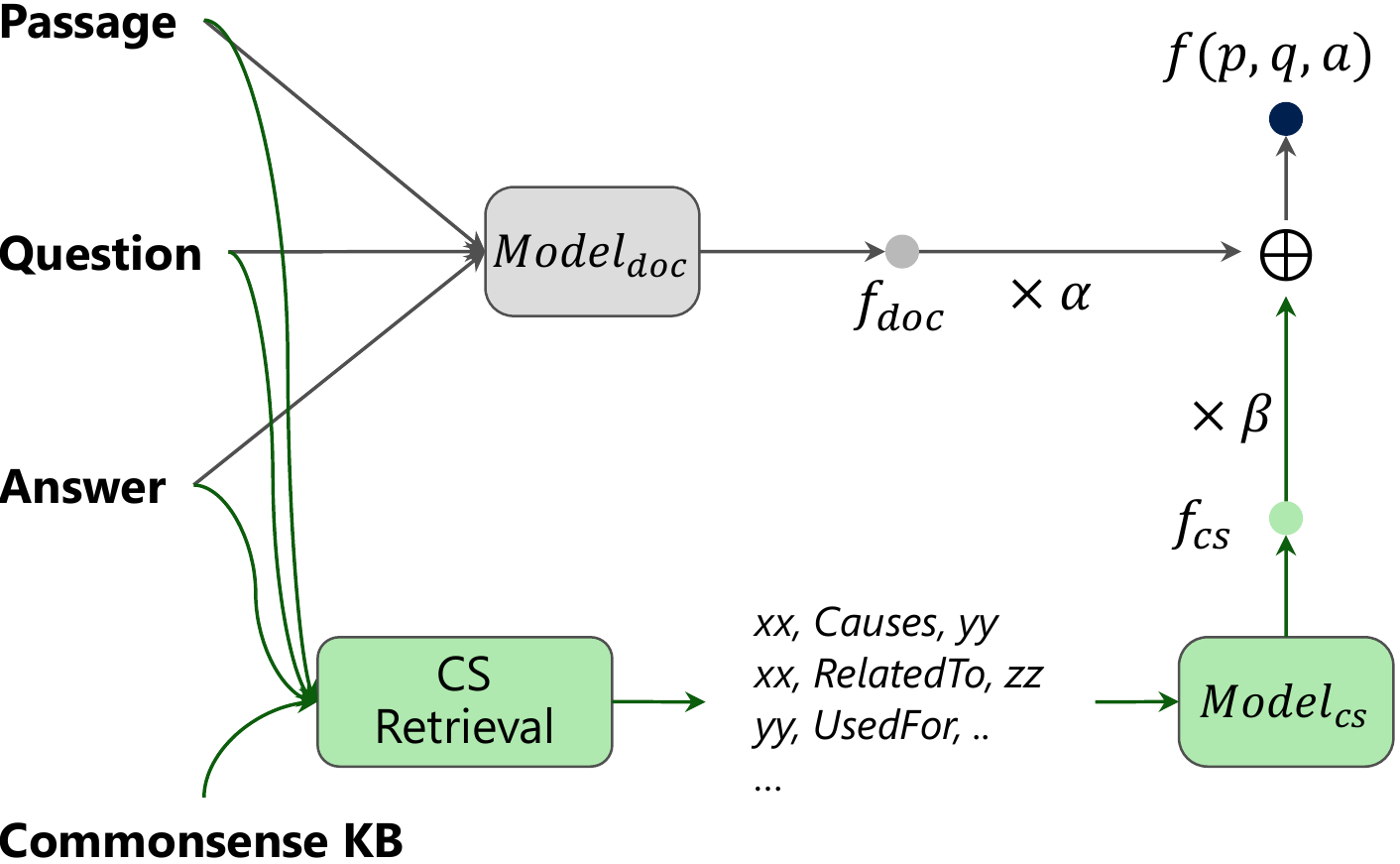}
	\caption{An overview of our system for commonsense based question answering.}
	\label{fig:model}
\end{figure}
An overview of the QA system is given in Figure \ref{fig:model}.
We define the scoring function $f(a_i)$ to calculate the score of a candidate answer $a_i$, which can be calculated by the sum of document based scoring function $f_{doc}(a_i)$ and commonsense based scoring function $f_{cs}(a_i)$
\begin{equation}
\label{equa:doc+cs}
f(a_i) = \alpha f_{doc}(a_i) + \beta f_{cs}(a_i)
\end{equation}
The calculation of the final score would consider the given passage, the given question, and a set of commonsense knowledge related to this instance.
In the next section we will detail the design and mathematical formulas of our commonsense knowledge based scoring function. 

\section{Commonsense-based Model}
In this section, we first describe how to pre-train commonsense-based functions to capture the semantic relationships between two concepts. 
Graph neural network \cite{scarselli2009graph} is used to integrate context from the graph structure in an external commonsense knowledge base.
Afterward, we present how to use the pre-trained functions to calculate the relevance score between two pieces of text, such as a question sentence and a candidate answer sentence.

\begin{figure}[h]
	\centering
	\includegraphics[width=.47\textwidth]{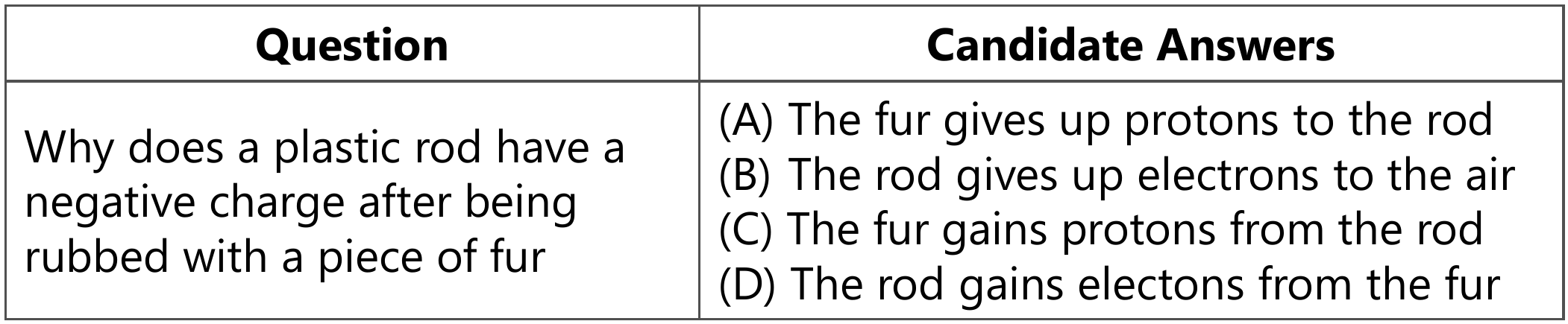}
	\caption{An example from ARC dataset. The analysis of this example could be improved if it is given the fact \{``\textit{electrons}'', ``\textit{HasA}'', ``\textit{negative charge}''\} in ConceptNet.}
	\label{fig:dc-example}
\end{figure}

We model both {direct} and {indirect}  relations between two concepts from commonsense KB, both of which are helpful when the connection between two sources (e.g., a question and a candidate answer) is missing based on the word utterances merely.
Take direction relation involved in Figure \ref{fig:dc-example} as an example. 
If a model is given the evidence from ConceptNet such that the concept ``\textit{electrons}'' and the concept ``\textit{negative charge}'' has direct relation, it would be more confident to distinguish between (B,D) and (A,C), thus has a larger probability of obtaining the correct answer (D). 
Therefore, it is desirable to model the relevance between the two concepts.
Moreover, ConceptNet could not cover all the concepts which potentially have direction relations. 
We need to model the direct relation for every two concepts.

Similarly, indirect relation also provides strong evidence for prediction making. 
As shown in the example of Fig \ref{fig:kb-sample}, the concept ``\textit{a car}''  has an indirect relation to the concept ``\textit{getting to a destination}'', both of  which have a direct connection to the pivot concept ``\textit{driving}''. 
With access to this information, a model would give a higher score to the answer containing ``\textit{car}'' when questioned ``\textit{how did someone get to the destination}''. 

Therefore, we model the commonsense-based relation between two concepts $c_1$ and $c_2$ as follows, where $\odot$ means element-wise multiplication, $Enc(c)$ stands for an encoder that represents a concept $c$ with a continuous vector.
\begin{equation}
f_{cs}^{}(c_1, c_2) = Enc(c_1) \odot Enc(c_2)
\end{equation}
Specifically, we represent a concept with two types of information, namely the words it contains and the neighbors connected to it in the structural knowledge graph. 
From the first aspect, since each concept might consist of a sequence of words, we encode it by a bidirectional LSTM 
over Glove word vectors \cite{pennington2014glove}, where the concatenation of hidden states at both ends is used as the representation. We denote it as $h^w(c)$.
From the second aspect, we represent each concept based on the representations of its neighbors and the relations that connect them. We get inspirations from graph neural network \cite{scarselli2009graph}. 
We regard a relation that connects two concepts as the compositional modifier to modify the meaning of the neighboring concept. 
Matrix-vector multiplication is used as the composition function.
We denote the neighbor-based representation of a concept $c$ as $h^{n}(c)$, which is calculated as follows, where $r(c,c')$ is the specific relation between two concepts, $NBR(c)$ stands for the set of neighbors of the concept $c$,  $W$ and $b$ are model parameters. 
\begin{equation}
h^{n}(c) = \sum_{c'\in NBR(c)} (W^{r(c,c')} h^w(c') + b^{r(c,c')})
\end{equation}
The final representation of a concept $c$ is the concatenation of both representations, namely $Enc(c) = [h^w(c);h^n(c)]$.
We use a ranking-based loss function to train the parameters.

In this equation, $c_1$ and $c_2$ form a positive instance, which means that they have a relationship with each other, while $c_1$ and $c'$ form a negative instance. $mgn$ is the margin with value of 0.1 in the experiment. 
We can easily learn two functions to model direct and indirect relations between two concepts by having different definitions of what a positive instance is, and accordingly using different strategies to sample the training instances. 
For the direct relation, we set those directly adjacent entities pairs in the knowledge graph as positive examples and randomly select entity pairs that have no direct relationship as negative examples.
For the indirect relation, we select entity pairs that have a common neighbor as a positive instance and randomly select an equal number of entities pairs that have no one-hop or two-hop connected relations as negative instances. 


In our experiment, we retrieve commonsense facts from ConceptNet \cite{speer2012representing}. 
As described above, each fact from ConceptNet can be represented as a triple, namely $c=(subject,relation,object) $. 
For each sentence (or paragraph), we retrieve a set of facts from ConceptNet. 
Specifically, we first extract a set of the n-grams from each sentence. We experiment with $\{1,2,3\}$-gram in our searching process, and then, we save the commonsense facts from ConceptNet which contain one of the extracted n-grams. 
We denote the facts for a sentence $s$ as $E_s$.

Suppose we have obtained commonsense facts for a question sentence and a candidate answer, respectively, let us denote the outputs as $E_1$ and $E_2$. 
We can calculate the final score by the following formula. The intuition is to select the most relevant concept of each concept in $E_1$, and then aggregate all these scores by average. 
\begin{equation}
f_{cs}(a_i) = \frac{1}{|E_1|}\sum_{x\in E_1}\max_{y\in E_2}(f_{cs}(x,y))
\end{equation}

Our method differs from TransE \cite{bordes2013translating} in three aspects. Firstly, the goals are different. The goal of TransE is to embed entities and predicates/relations into low-dimensional vector space. Secondly, the outputs are different. TransE outputs embeddings of entities and predicates, while out model outputs the parameterized scoring function. Thirdly, the evidence used for representing entities are different. Compared to TransE, our model further incorporates the neighbors of concepts via graph neural network.

\section{Experiment}
We conduct experiments on three question answering datasets, namely SemEval 2018 Task 11 \cite{ostermann2018semeval}, ARC Challenge Dataset \cite{clark2018think} and OpenBook QA Dataset \cite{mihaylov2018can} to evaluate the effectiveness of our system. To improve the generality of our model, we trained the document based model and commonsense based model separately, which can make the commonsense based model easier to be incorporated into other tasks.
On ARC, SemEval and OpenBook QA datasets, we follow existing studies and use accuracy as the evaluation metric. 


\begin{figure*}[t]
	\centering
	\includegraphics[width=\textwidth]{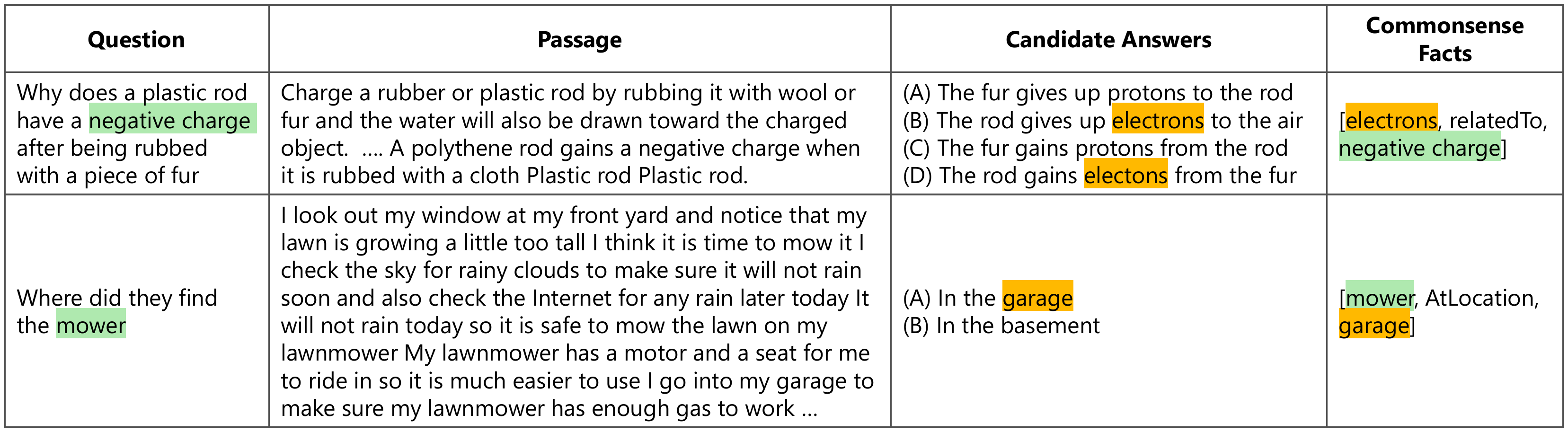}
	\caption{Examples that require commonsense-based direct relations between concepts on ARC and SemEval datasets.}
	\label{fig:output-dir}
\end{figure*}
\subsection{Model Comparisons and Analysis}

\begin{table}[h]\small
	\centering
	\begin{tabular}{l|c}
		\hline
		Model & Accuracy \\
		\hline
		IR \cite{clark2018think} & 20.26\% \\
		TupleInference \cite{clark2018think}  & 23.83\% \\
		DecompAttn \cite{clark2018think}  & 24.34\% \\
		Guess-all \cite{clark2018think}  & 25.02\% \\
		DGEM-OpenIE \cite{clark2018think}  & 26.41\% \\
		BiDAF \cite{clark2018think}  & 26.54\% \\
		Table ILP \cite{clark2018think}  & 26.97\% \\
		DGEM \cite{clark2018think}  & 27.11\% \\
		KG$^2$ \cite{zhang2018kg} & 31.70\%\\
		BiLSTM Max-out \cite{Mihaylov2018CanAS} & 33.87\% \\
		ET-RR \cite{Ni2018LearningTA} & 36.36\% \\
		Reading Strategies \cite{sun2018improving} & 42.32\% \\
		\hline
		TriAN  & 31.25\% \\
		TriAN + $f_{cs}^{dir}$  & 32.28\% \\
		TriAN + $f_{cs}^{ind}$   & 32.96\% \\
		TriAN + $f_{cs}^{dir}$ + $f_{cs}^{ind}$& 33.39\% \\\hline
		TriAN(Concat Bert) & 35.18\% \\
		TriAN(Concat Bert)+$f_{cs}^{dir}$ + $f_{cs}^{ind}$ & 36.55\% \\\hline
	\end{tabular}
	\caption{Performances of different approaches on the ARC Challenge dataset. 
	}
	\label{table:result of ARC}
\end{table}

Table \ref{table:result of ARC}, Table \ref{table:result of OpenBook QA} and Table \ref{table:result of SemEval} show the results on these three datasets, respectively.
On the ARC and OpenBook QA dataset, we compare our model with a list of existing \mbox{systems}. 
On the SemEval dataset, we only report the results of TriAN, which is the top-performing system in the SemEval evaluation\footnote{During the SemEval evaluation, systems including TriAN report results based on model pretraining on RACE dataset \cite{lai2017race} and system ensemble. In this work, we report numbers on SemEval without pre-trained on RACE or ensemble.}.
$f_{cs}^{dir}$ is our commonsense-based model for direct relations, and $f_{cs}^{ind}$ represents the commonsense-based model for indirect relations. 
We can observe that commonsense-based scores improve the accuracy of the document-based model TriAN, and combining both scores could achieve further improvements on both datasets. The results show that our commonsense-based models are complementary to standard document-based models. We also apply BERT \cite{devlin2018bert} to improve our baseline and show our method enhance the performance on the stronger baseline. 

\begin{table}[h]\small
	\centering
	\begin{tabular}{l|c}
		\hline
		Model & Accuracy \\
		\hline
		\multicolumn{2}{l}{NO TRAINING, $F$+KB}\\
		\hline
		IR \cite{mihaylov2018can} & 24.8\% \\
		TupleInference \cite{mihaylov2018can}  & 26.6\% \\
		DGEM \cite{mihaylov2018can}  & 24.6\% \\
		PMI \cite{mihaylov2018can} & 21.2\% \\
		\hline
		\multicolumn{2}{l}{TRAINED MODELS, NO $F$ or KB}\\
		\hline
		Embedd+Sim \cite{mihaylov2018can}  & 41.8\% \\
		ESIM \cite{mihaylov2018can}  & 48.9\% \\
		PAD \cite{mihaylov2018can}  & 49.6\% \\
		Odd-one-out Solver \cite{mihaylov2018can}  & 50.2\% \\
		Question Match \cite{mihaylov2018can}  & 50.2\% \\
		Reading Strategies \cite{sun2018improving} & 55.8\% \\
		\hline
		\multicolumn{2}{l}{ORACLE MODELS, $F$ AND/OR KB}\\
		\hline
		$f$ \cite{mihaylov2018can} & 55.8\%\\
		$f$ + WordNet \cite{mihaylov2018can} & 56.3 \%\\ 
		$f$ + ConceptNet \cite{mihaylov2018can} & 53.7 \%\\ 
		\hline
		TriAN  & 56.6\% \\
		TriAN + $f_{cs}^{dir}$ + $f_{cs}^{ind}$& 58.0\% \\\hline
		TriAN + BERT & 70.6\% \\
		TriAN + BERT+ $f_{cs}^{dir}$ + $f_{cs}^{ind}$& 72.8\% \\
		\hline
	\end{tabular}
	\caption{Performances of different approaches on the OpenBook QA dataset. $F$ indicates the golden fact for the question.
	}
	\label{table:result of OpenBook QA}
\end{table}

We give examples from ARC and SemEval datasets that are incorrectly predicted by the document-based model, while correctly solved by incorporating the commonsense-based models.
Figure \ref{fig:output-dir} shows two examples that require commonsense-based direct relations between concepts. 
The first example comes from the ARC. We can see that the retrieved facts from ConceptNet provide useful evidence to connect question to candidate answers (B) and (D). By combining with the document-based model, which might favor candidates with the co-occurred word ``\textit{fur}'', the final system might give a higher score to (D).
The second example is from SemEval. Similarly, we can see that the retrieved facts from ConceptNet are helpful in making the correct prediction.
\begin{table}[h]\small
	\centering
	\begin{tabular}{l|c}
		\hline
		Model &Accuracy\\
		\hline
		HMA \cite{chen2018hfl} & 80.9\% \\
		Reading Strategies \cite{sun2018improving} & 88.8\% \\
		\hline
		TriAN & 80.33\% \\
		TriAN + $f_{cs}^{dir}$ & 81.58\% \\
		TriAN + $f_{cs}^{ind}$ & 81.44\% \\
		TriAN + $f_{cs}^{dir}$ + $f_{cs}^{ind}$& 81.80\% \\\hline
		TriAN + BERT & 86.27\% \\
		TriAN + BERT+ $f_{cs}^{dir}$ + $f_{cs}^{ind}$& 87.49\% \\
		\hline
		
	\end{tabular}
	\caption{Performances of different approaches on the SemEval Challenge dataset. 
	}
	\label{table:result of SemEval}
\end{table}

\begin{figure*}[t]
	\centering
	\includegraphics[width=\textwidth]{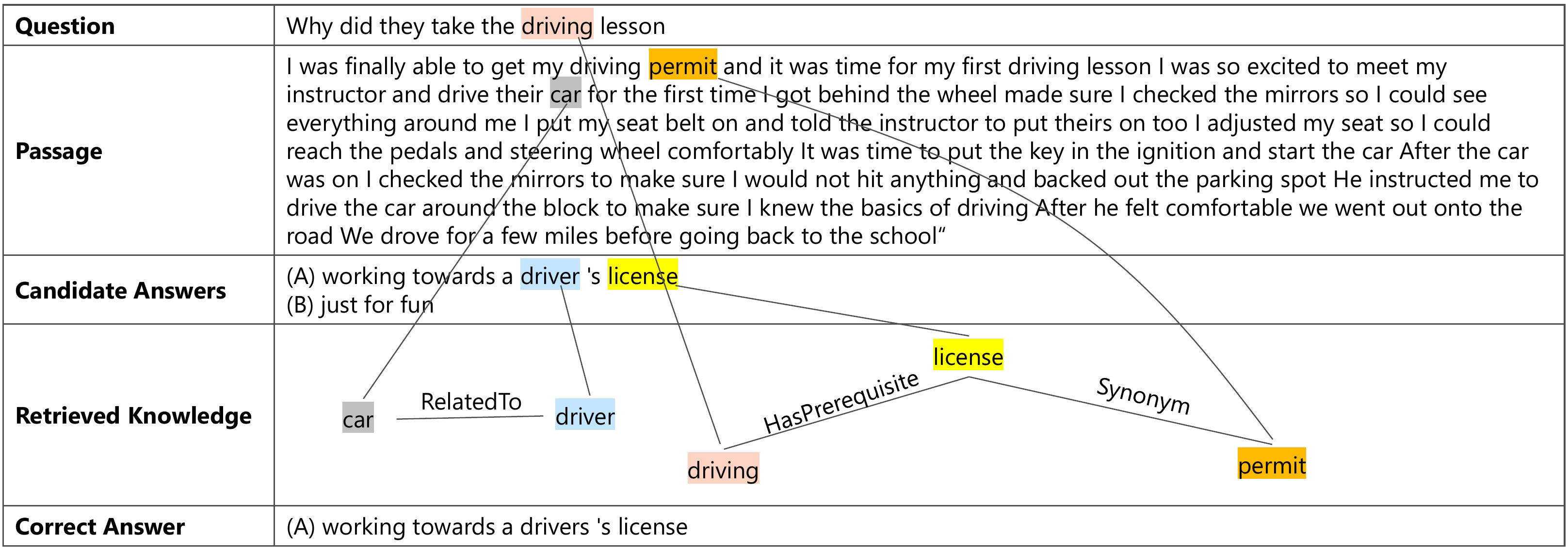}
	\caption{An example from SemEval 2018 that requires sophistic reasoning based on commonsense knowledge.}
	\label{fig:cs-sample}
\end{figure*}

Figure \ref{fig:cs-sample} shows an example from SemEval that benefits from both direct and indirect relations from commonsense knowledge. Despite both the question and candidate (A) mention about ``\textit{drive/driving}'', the document-based model fails to make the correct prediction.
We can see that the retrieved facts from ConceptNet help from different perspectives. The fact \{``\textit{driving}",``\textit{HasPrerequisite}",``\textit{license}"\} directly connects the question to the candidate (A), and  both   \{``\textit{license}",``\textit{Synonym}",``\textit{permit}"\} and \{``\textit{driver}",``\textit{RelatedTo}",``\textit{care}"\} directly connects candidate (A) to the passage. 
Besides, we calculate for the question-passage pair, where the indirect relation between \{``\textit{driving}",``\textit{permit}"\} could be used as side information for prediction.
We further make comparisons by implementing different strategies to use commonsense knowledge from ConceptNet. 
We implement three baselines, including \textbf{TransE} \cite{bordes2013translating}, Pointwise Mutual Information (\textbf{PMI}) and Key-Value Memory Network (\textbf{KV-MemNet}) \cite{DBLP:journals/corr/MillerFDKBW16}. 
\begin{table}[h]\small
	\centering
	\begin{tabular}{l|ccc}
		\hline
		Model &ARC& SemEval &OBQA\\
		\hline
		TriAN  & 31.25\% & 80.33\%&56.6\%\\
		TriAN + PMI  & 31.72\% & 80.50\% & 53.1\%\\
		TriAN + TransE  & 30.59\% & 80.37\% & 55.2\%\\
		TriAN + KV-MemNet  & 30.49\% & 80.59\% & 54.6\%\\
		TriAN + $f_{cs}^{dir}$ + $f_{cs}^{ind}$ & 33.39\% & 81.80\%& 58.0\%\\\hline
	\end{tabular}
	\caption{Performances of approaches with different strategies to use commonsense knowledge on ARC, SemEval 2018 Task 11 and OpenBook QA datasets.
	}
	\label{table:results-diff-cs}
\end{table}
From Table \ref{table:results-diff-cs} we can see that learning direct and indirection connections based on contexts from word-level constituents and neighbor from knowledge graph performs better than TransE which is originally designed for KB completion. PMI performs well, however, its performance is limited by the information it can take into account, i.e. the word count information. 
The comparison between KV-MemNet and our approach further reveals the effectiveness of pretraining.

\subsection{Error Analysis and Discussion}
We analyze the wrongly predicted instances and summarize the majority of errors of the following groups.

The {first} type of error, which is also the dominant one, is caused by failing to highlight the most useful concept in all the retrieved ones.
The usefulness of a concept should also be measured by its relevance to the question, its relevance to the document, and whether introducing it could help distinguish between candidate answers.
For example, the question is ``\textit{Where was the table set}'' is asked based on a document talking about dinner, according to which two candidate answers are ``\textit{On the coffee table}'' and ``\textit{At their house}''. Although the retrieved concepts for the first candidate answer also being relevant, they are not relevant to the question type ``\textit{where}''.  
The problem would be alleviated by incorporating a context-aware module to model the importance of a retrieved concept in a particular instance and combining it with the pre-trained model to make the final prediction.

The second type of error is caused by the ambiguity of the entity/concept to be linked to the external knowledge base. 
For example, suppose the document talks about computer science and machine learning, the concept ``\textit{Micheal Jordan}'' in question should be linked to the machine learning expert rather than the basketball player. However, to achieve this requires an entity/concept disambiguation model, the input of which also considers the question and the passage.

Moreover, the current system fails to handle difficult questions which need logical reasoning, such as ``\textit{How long do the eggs cook for}'' and ``\textit{How many people went to the movie together}''. 
We believe that deep question understanding, such as parsing a question based on a predefined grammar and operators in a semantic parsing manner, is required to handle these questions, which is a promising direction, and we leave it to future work.
\section{Related Work}

Our commonsense-based model, which is pre-trained on commonsense KB, 
relates to recent neural network approaches that incorporate side information from external and structured knowledge bases.
Existing studies roughly fall into two groups, where the first group aims to enhance each basic computational unit 
 and the second group aims to support external signals at the top layer before the model makes the final decision.
The majority of works fall into the first group. For example, \newcite{yang2017leveraging} use concepts from WordNet and NELL, and weighted average vectors of the retrieved concepts to calculate a new LSTM state. \newcite{mihaylov2018knowledgeable} retrieve relevant concepts from external knowledge for each token, and get an additional vector with a solution similar to the key-value memory network. 
\cite{khot2017answering}, text entailment \cite{chen2018neural},
 etc. We believe that this line might work well on a specific dataset; however, the model only learns overlapped knowledge between the task-specific data and the external knowledge base. Thus, the model may not be easily adapted to another task/dataset where the overlapped is different from the current one.
Our work belongs to the second group. \newcite{lin2017reasoning} learn the correlation between concepts with pointwise mutual information. We explore richer contexts from the rational knowledge graph with the graph-based neural network and empirically show that the approach performs better on question answering datasets.

Our work relates to the field of model pretraining in NLP and computer vision fields \cite{mahajan2018exploring}. In the NLP community, works on model pretraining can be divided into unstructured text-based and structured knowledge-based ones.
Both word embedding learning algorithms \cite{pennington2014glove} and contextual embedding learning algorithms \cite{peters2018deep,devlin2018bert} belong to the text-based direction. Compared with these methods, which aim to learn a representation for a continuous sequence of words, our goal is to model the concept relatedness with graph structure in the knowledge base.
Previous works on knowledge-based pretraining are typically validated on knowledge base completion or link prediction task \cite{bordes2013translating,socher2013reasoning}. 
Our work belongs to the second line. We pre-train models from the commonsense knowledge base and apply the approach to the question answering task. 
We believe that combining both structured knowledge graphs and unstructured texts to do model pretraining is very attractive, and we leave this for future work.

\section{Conclusion}
We work on commonsense based question answering tasks. We present a simple and effective way to pre-train models to measure relations between concepts. Each concept is represented based on its internal information (i.e., the words it contains) and external context (i.e., neighbors in the knowledge graph).
We use ConceptNet as the external commonsense knowledge base, and apply the pre-trained model on three question answering tasks (ARC, SemEval and OpenBook QA).
Results show that the pre-trained models are complementary to standard document-based neural network approaches and could make further improvement through model combination. 
Model analysis shows that our system could discover useful evidence from commonsense knowledge base.
In the future, we plan to address the issues raised in the discussion part including incorporating a context-aware module for concept ranking and considering logical reasoning operations.
We also plan to apply the approach to other datasets that require commonsense reasoning.


\bibliographystyle{named}
\bibliography{ijcai19}

\end{document}